\nonstopmode
\pdfoutput=1

\documentclass[11pt]{article}

\usepackage{emnlp2021}
\usepackage{times}
\usepackage{latexsym}
\usepackage{amssymb}
\usepackage{amsmath}
\usepackage{booktabs}
\usepackage{color,soul}
\usepackage{multirow}
\usepackage{float}
\usepackage{graphicx}
\usepackage[ruled,vlined]{algorithm2e}
\usepackage{times}
\usepackage{latexsym}

\usepackage[T1]{fontenc}

\usepackage[utf8]{inputenc}
\graphicspath{{figures/}}
\usepackage{microtype}

\newcommand\benchmark{\textsc{AfroMT}}
\newcommand{\afrobart}{AfroBART}
\newcommand{\afrobartdict}{AfroBART-Dictionary}
\newcommand{\afrobartbaseline}{AfroBART Baseline}

\usepackage{amsmath,amsfonts,bm}

\def\eqref#1{equation~\ref{#1}}

\def\1{\bm{1}}

\def\mM{{\bm{M}}}

\DeclareMathAlphabet{\mathsfit}{\encodingdefault}{\sfdefault}{m}{sl}
\SetMathAlphabet{\mathsfit}{bold}{\encodingdefault}{\sfdefault}{bx}{n}

\def\gD{{\mathcal{D}}}

\newcommand{\insertappendix}{
\newpage~\newpage
\appendix
\section{\benchmark}
We provide extra information --- Script, Language Family, L1 and L2 speakers, Location as well as Word Order --- in Table~\ref{tab:extra}.

We upload \benchmark~as well as the data generated using the pseudo monolingual data synthesis\footnote{Note that we do not generate pseudo monolingual data for Afrikaans due to its high similarity with Dutch --- a high resource language.}.

\section{Pretraining}
\paragraph{Data} We use in addition to the monolingual data for the languages in \benchmark~(shown in Table 1 of the main paper), we 14 GB of English data, and 7 GB of French and Dutch data each.
\paragraph{Additional Hyperparameters} We optimize the model using Adam \citep{kingma2017adam} using hyperparameters $\beta=(0.9,0.98)$ and $\epsilon=10^{-6}$. We warm up the learning rate to a peak of $3 \times 10^{-4}$ over 10K iterations and then decay said learning rate using the polynomial schedule for 90K iterations. For regularization, we use a dropout value of $0.1$ and weight decay of $0.01$. 

\section{Finetuning Hyperparameters}
\paragraph{Training from scratch} When training using random initialization (or CLT), we use a batch size of 32K (or 64K in the case of CLT) tokens and warmup the learning rate to $5 \times 10^{-4}$ over 10K iterations and decay with the inverse square root schedule. We use a dropout value of $0.3$, a weight decay value of $0.01$, and a label smoothing value of $\epsilon=0.1$.
\paragraph{Finetuning from \afrobart} We train using a batch size of 32K tokens, and use a smaller learning rate of $3 \times 10^{-4}$. We use a polynomial learning rate schedule, maximizing the learning rate at 5000 iterations and finishing training after 50K iterations. We perform early stopping, stopping training if the best validation loss remains constant for over 10 epochs.  We use a label smoothing value of $\epsilon=0.2$, a dropout value of $0.3$ and weight decay of $0.01$.

\section{Training Infrastructure}
For finetuning models on \benchmark~we use between 1 and 8 NVIDIA V100 16GB GPUs on a DGX-1 machine running Ubuntu 16.04 on a Dual 20-Core Intel Xeon E5-2698 v4 2.2 GHz. For pretraining we make use of a compute cluster using 8 nodes with 4 NVIDIA V100 16GB GPUs per node.

\section{Quantification of Potential Data Leakage}
In low-resource machine translation, data-leakage is a key concern given its pertinence in the mitigation of misleading results. We quantify data leakage for our benchmark We measured the target-side train-test data leakage using the 4-gram overlap between the training/test sets. We take the most frequent 100k 4-grams from the training set and compare them with all 4-grams in the test set and obtain an average 4-gram overlap of 5.01$\pm$2.56\% (measured against all test-set 4-grams). To put this value in context, we ran these on other widely used low-resource datasets from IWSLT (En-Vi, Ja-En, Ar-En) and obtained 9.50\%, 5.49\%, and 5.53\% respectively. We believe this to be reasonable evidence of the lack of train-test data-leakage.

Furthermore, we also show improvements on source-target leakage as follows: we compute BLEU between the source and target over all training sets, we obtain an average of 4.5$\pm$1.3 before cleaning (indicating heavy overlap in certain source-target pairs in the corpus), and after cleaning 0.7$\pm$0.2 indicating a significant decrease in such overlap.

\section{Parameter Count}
We keep the parameter count of 85M consistent throughout our experiments as we use the same model architecture. We ran experiments on scaling up randomly initialized models with a hidden size of 768 and feed forward dimension of 3072 with 6 layers in both the encoder and decoder on three language pairs. The results of these experiments can be seen in Table~\ref{tab:my_label}.
\begin{table}[H]
    \centering
    \footnotesize
    \scalebox{0.95}{\begin{tabular}{llccc}
    \toprule[0.13em]
    \textbf{Lang. Pair} & \textbf{Model} & \textbf{Param. Count} & \textbf{BLEU} & \textbf{chrF}\\
    \midrule[0.07em]
        \multirow{2}{*}{En-Run} & Random & 85M  & 22.92 & 51.89 \\
         & Random & 160M & 22.12 & 51.22\\
         \midrule[0.07em]
        \multirow{2}{*}{En-Sw} & Random & 85M  & 33.61 & 58.56 \\
         & Random & 160M & 33.62 & 58.65\\
                  \midrule[0.07em]
        \multirow{2}{*}{En-Ln} & Random & 85M  & 28.37 & 53.65 \\
         & Random & 160M & 27.58 &  53.29 \\
    \bottomrule[0.13em]
    \end{tabular}}
    \caption{Scalability comparison
    }
    \label{tab:my_label}
\end{table}
\begin{table*}
\centering
\begin{tabular}{cccccc}
\toprule[0.15em]

\multirow{2}{*}{ \textbf{Languages} } & \multirow{2}{*}{ \textbf{ISO 639-2 code} } & \multirow{2}{*}{ \textbf{Script} } & \multirow{2}{*}{ \textbf{Language Family} } & \multicolumn{2}{c}{ \textbf{Population} } \\
& & & & \textbf{L1} &\textbf{L2} \\
\midrule[0.07em]
Afrikaans & Afr & Latin, Arabic & Indo-European: Germanic & 7.2M & 10.3M \\
Bemba & Bem & Latin & Niger-Congo: Bantu Zone M & 4M & 2M \\
Lingala & Lin & Latin & Niger-Congo: Bantu Zone C & 20M & 25M \\
Rundi & Run & Latin & Niger-Congo: Bantu Zone D & 11.9M & --- \\
Sotho & Sot & Latin & Niger-Congo: Bantu Zone S & 5.6M & 7.9M \\
Swahili & Swa & Latin & Niger-Congo: Bantu Zone G & 150M & 90M \\
Xhosa & Xho & Latin & Niger-Congo: Bantu Zone S & 8.2M & 11M \\
Zulu & Zul & Latin & Niger-Congo: Bantu Zone S & 12M & 16M \\
\bottomrule[0.15em]
\end{tabular}
\begin{tabular}{cccccc}
\toprule[0.15em]
\multirow{2}{*}{ \textbf{Languages} } & \multirow{2}{*}{ \textbf{Location} } & \textbf{Noun Classes} & \multirow{2}{*}{ \textbf{Word Order} } \\
& & Singular/Plural/Total & \\
\midrule[0.07em]
Afrikaans & South Africa, Namibia & --- & SVO \\
Bemba & North-Eastern Zambia & 9/6/15 & SVO \\
Lingala & DR Congo, Congo & 9/6/15 & SVO \\
Rundi & Burundi & 9/7/16 & SVO \\
Sotho & Lesotho, South Africa, Zimbabwe & 6/5/11 & SVO \\
Swahili & \small{African Great Lakes region, East/Southern Africa} & 9/9/18 & SVO \\
Xhosa & South Africa & 8/7/15 & SVO \\
Zulu & South Africa, Lesotho, Eswatini & 6/10/16 & SVO \\
\bottomrule[0.15em]
\end{tabular}
\caption{Extra information on all the languages contained within \benchmark} \label{tab:extra}
\end{table*}
It can be seen that increasing parameter count of AfroMT for random initialization doesn't provide
an effective performance/compute tradeoff, harming performance on English-Rundi and English-Lingala, while minmally improving performance on English-Swahili. This being said, we believe that if we scale up \afrobart, given the insights from \citet{liu2020mbart,gordon2021data}, we can provide a good initialization to allows us to scale to these model sizes for greater performance.

\section{Fine-grained morphological analysis in a data constrained regime}
\begin{figure}[H]
    \centering
    \includegraphics[width=\linewidth]{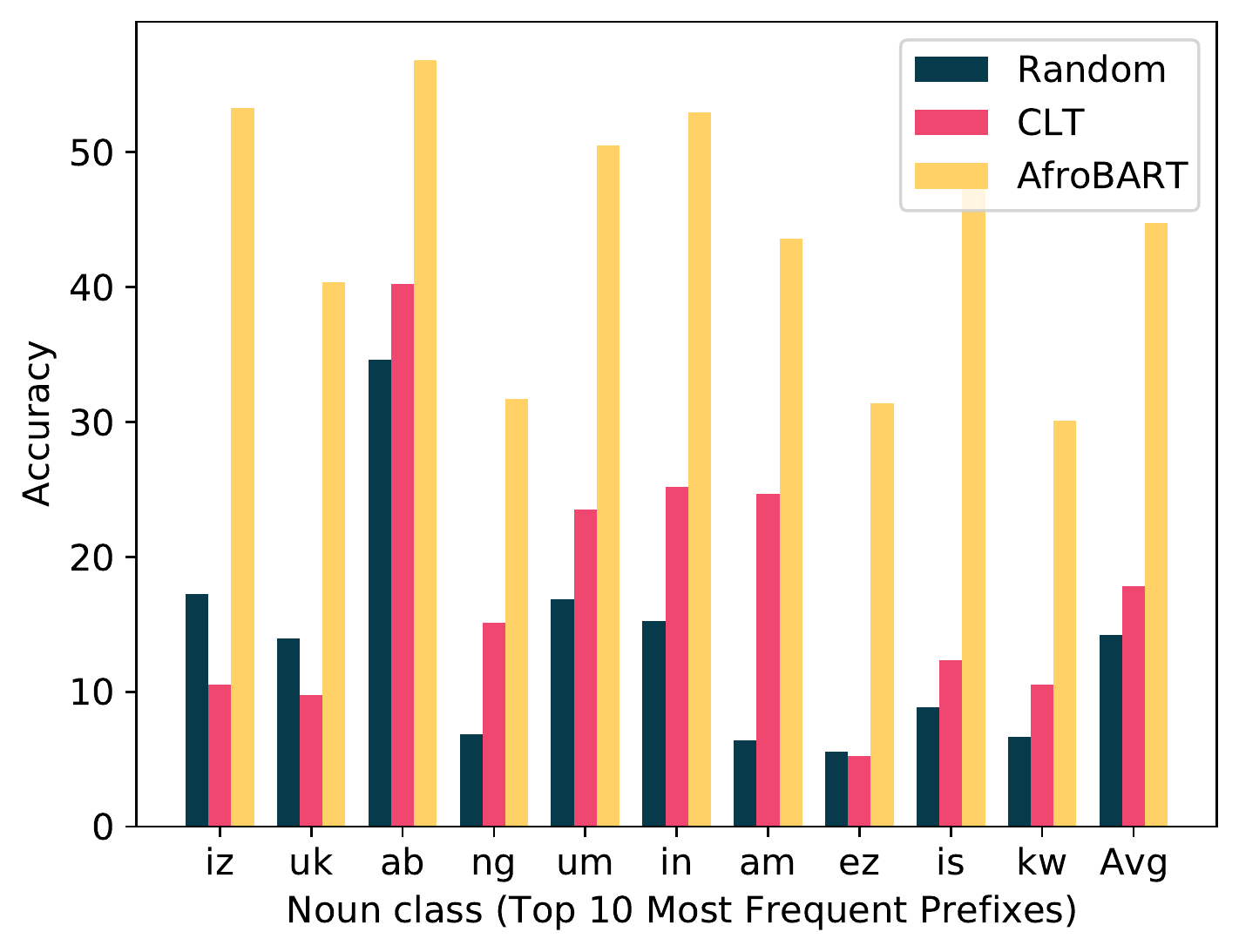}
    \caption{Translation accuracy of the AfroBART and Random baseline systems on Zulu (10k pairs) noun classes with top 10 most frequent 2-character prefixes. }
    \label{fig:zulu_acc}
\end{figure}
We perform our fine grained morphological analaysis (described in Section \S5.3 of the main paper) on the data constrained scenario (described in Section \S5.2 of the main paper).  We perform the analysis on English-Xhosa and English-Zulu (10k parallel sentence pairs) side by side and visualize them in Figure \ref{fig:zulu_acc} and Figure \ref{fig:xhosa_acc}. It can be seen that cross lingual transfer improves accuracy in this data constrained scenario over a random baseline, which is inturn improved upon by \afrobart.
\begin{figure}[H]
    \centering
    \includegraphics[width=0.9\linewidth]{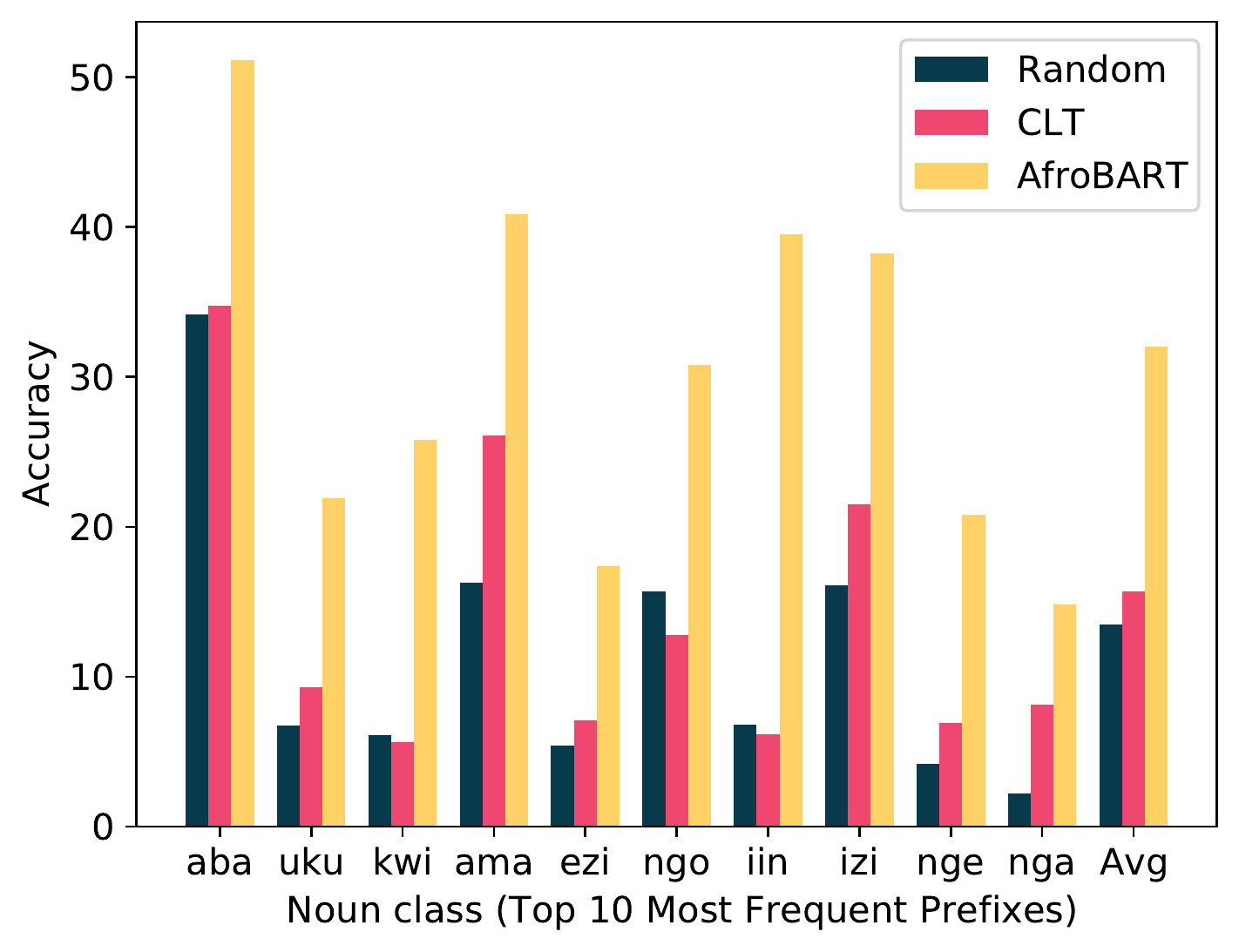}
    \caption{Translation accuracy of the AfroBART and Random baseline systems on Xhosa (10k pairs) noun classes with top 10 most frequent 3-character prefixes. (Same as Figure 5 of the main paper)}
    \label{fig:xhosa_acc}
\end{figure}
Additionally, we report the BLEU and chrF scores of the data constrained experiments (shown in Figure 3 of the main paper) in Table~\ref{tb:small_data}.
\begin{table}[H]
    \centering
    \footnotesize 
    \begin{tabular}{llccc}
    \toprule[0.13em]
    \textbf{Lang. Pair} & \textbf{\# Data}& \textbf{Model}  & \textbf{BLEU} & \textbf{chrF}\\
    \midrule[0.07em]
        \multirow{9}{*}{En-Zu} & \multirow{3}{*}{10k}   & Random& 4.06 & 28.26 \\
        & & CLT & 8.08 & 37.9 \\
        & & \afrobart & \textbf{20.44} &  \textbf{51.35} \\
        \cmidrule(lr){2-5}
        & \multirow{3}{*}{50k}   & Random& 18.01 & 50.55 \\
        & & CLT & 20.41 & 51.52 \\
        & & \afrobart & \textbf{26.95} &  \textbf{58.56} \\
        \cmidrule(lr){2-5}
        & \multirow{3}{*}{100k}   & Random& 23.09 & 55.63 \\
        & & CLT & 24.50 & 55.81 \\
        & & \afrobart & \textbf{29.41} & \textbf{60.81} \\
    \midrule[0.07em]
        \multirow{9}{*}{En-Xh} & \multirow{3}{*}{10k}   & Random& 2.82 & 26.29 \\
        & & CLT &  6.35 & 32.31 \\
        & & \afrobart & \textbf{13.98}  &  \textbf{43.19} \\
        \cmidrule(lr){2-5}
        & \multirow{3}{*}{50k}   & Random& 11.94 & 42.62 \\
        & & CLT & 10.12  & 39.73 \\
        & & \afrobart & \textbf{18.54}  &  \textbf{49.70} \\
        \cmidrule(lr){2-5}
        & \multirow{3}{*}{100k}   & Random& 16.00 & 47.92 \\
        & & CLT &  11.64 & 41.19 \\
        & & \afrobart & \textbf{20.45}  & \textbf{ 52.35} \\
    \bottomrule[0.13em]
    \end{tabular}
    \caption{Comparing performance with various amounts of parallel data} %
    \label{tb:small_data}
\end{table}}

\title{{\benchmark}: Pretraining Strategies and Reproducible Benchmarks for Translation of 8 African Languages}

\author{Machel Reid$^1$, Junjie Hu$^2$, Graham Neubig$^2$, Yutaka Matsuo$^1$\\
  $^1$The University of Tokyo, $^2$Carnegie Mellon University \\
  \texttt{\{machelreid,matsuo\}@weblab.t.u-tokyo.ac.jp}\\
  \texttt{\{junjieh,gneubig\}@cs.cmu.edu}
 }
 
\begin{document}
\maketitle
\begin{abstract}

Reproducible benchmarks are crucial in driving progress of machine translation research.
However, existing machine translation benchmarks have been mostly limited to high-resource or well-represented languages. Despite an increasing interest in low-resource machine translation, there are no standardized reproducible benchmarks for many African languages, many of which are used by millions of speakers but have less digitized textual data.
To tackle these challenges, we propose \benchmark, a standardized, clean, and reproducible machine translation benchmark for eight widely spoken African languages. We also develop a suite of analysis tools for system diagnosis taking into account unique properties of these languages. Furthermore, we explore the newly considered case of \textit{low-resource focused pretraining} and develop two novel data augmentation-based strategies, leveraging word-level alignment information and pseudo-monolingual data for pretraining multilingual sequence-to-sequence models. We demonstrate significant improvements when pretraining on 11 
languages, with gains of up to 2 BLEU points over strong baselines. We also show gains of up to 12 BLEU points over cross-lingual transfer baselines in data-constrained scenarios.
All code and pretrained models will be released as further steps towards larger reproducible benchmarks for African languages.\footnote{Source code, pretrained models, and data can be found at \url{https://github.com/machelreid/afromt}}

\end{abstract}

\section{Introduction}

Accuracy of machine translation systems in many languages has improved greatly over the past several years due to the introduction of neural machine translation (NMT) techniques \citep{bahdanau2014eural,sutskever2014equence,vaswani2017attention}, as well as scaling to larger models \citep{ott-etal-2018-scaling}.
However, many of these advances have been demonstrated in settings where very large parallel datasets are available \citep{meng2019large,arivazhagan2019massively}, and NMT systems often underperform in low-resource settings when given small amounts of parallel corpora \citep{koehn-knowles-2017-six,guzman2019he}.
One solution to this has been leveraging multilingual pretraining on large sets of monolingual data \citep{lample2019crosslingual,song2019mass,liu2020mbart}, leading to improvements even with smaller parallel corpora.
However, this thread of work has focused on scenarios with the following two properties: (1) pretraining on a plurality of European languages
and (2) cases in which the monolingual pretraining data greatly exceeds the parallel data used for finetuning (often by over 100 times) \citep{guzman2019he,liu2020mbart}.

However, in the case of many languages in the world, the above two properties are often not satisfied.
In particular, taking the example of African languages (the focus of our work), existing (small) parallel corpora for English-to-African language pairs often comprise the majority of available monolingual data in the corresponding African languages.
In addition, African languages are often morphologically rich and from completely different language families, being quite distant
from European languages. Moreover, despite the importance of reproducible benchmarks to measuring progress on various tasks in an empirical setting, there exists no standardized machine translation benchmark for the majority of African languages.

In this work, we introduce (1) a new machine translation benchmark for African languages, and (2) pretraining techniques to deal with the previously unexplored case where the size of monolingual data resources for pretraining is similar or equal to the size of parallel data resources for fine-tuning, and (3) evaluation tools designed for measuring qualities regarding the unique grammar of these languages in machine translation systems for better system evaluation. 

Our proposed benchmark, \benchmark, consists of translation tasks between English and 8 African languages --- Afrikaans, Xhosa, Zulu, Rundi, Sesotho, Swahili, Bemba, and Lingala --- four of which are not included in commercial translation systems such as Google Translate (as of Feb.~2021). In \S\ref{sec:benchmark}, we describe the detailed design of our benchmark, including the language selection criterion and the methodology to collect, clean and normalize the data for training and evaluation purposes. In \S\ref{sec:methods}, we provide a set of strong baselines for our benchmark, including denoising sequence-to-sequence pretraining \citep{lewis-etal-2020-bart,liu2020mbart}, transfer learning with similar languages~\citep{zoph-etal-2016-transfer,neubig-hu-2018-rapid}, and our proposed data augmentation methods for pretraining on low-resource languages. Our first method leverages bilingual dictionaries to augment data in high-resource languages (HRL), and our second method iteratively creates pseudo-monolingual data in low-resource languages (LRL) for pretraining. Extensive experiments in \S\ref{sec:experiments} show that our proposed methods outperform our baselines by up to $\sim$2 BLEU points over all language pairs and up to $\sim$15 BLEU points in data-constrained scenarios.

\section{\benchmark~benchmark} \label{sec:benchmark}
In this section, we detail the construction of our new benchmark, \benchmark. We first introduce our criteria for selecting the languages (\S\ref{sec:criteria}), and then describe the steps to prepare the dataset (\S\ref{sec:datasources}, \ref{sec:preparation}).

\subsection{Language Selection Criteria} \label{sec:criteria}

Given \benchmark's goal of providing a reproducible evaluation of African language translation, we select languages based on the following criteria: 
\paragraph{Coverage of Speakers \& Language Representation} We select languages largely based on the coverage of speakers as well as how represented they are in commercial translation systems. In total, the \benchmark~benchmark covers 225 million L1 and L2 speakers combined, covering a large number of speakers within Sub-Saharan Africa.

\paragraph{Linguistic Characteristics}
With the exception of English and Afrikaans, which belong to the Indo-European language family, all of the considered languages belong to the Niger-Congo family which is Africa's largest language family in terms of geographical area and speaking population (see Appendix). Similar to English, the Niger-Congo family generally follows the SVO word order. One particular characteristic feature of these languages is their morphosyntax, especially their system of noun classification, with noun classes often exceeding 10, ranging from  markers denoting male/female/animate/inanimate and more\footnote{\url{https://en.wikipedia.org/wiki/Niger\%E2\%80\%93Congo_languages}}. These noun classes can be likened in some sense to the male/female designation found in romance languages. However, in contrast with these languages, noun markers in Niger-Congo languages are often integrated within the word, usually as a prefix \citep{bendorsamuel1989niger}. For example: in Zulu, \textit{isiZulu} refers to the \textit{Zulu language}, whereas \textit{amaZulu} refers to the \textit{Zulu people}. Additionally, these languages also use~``verb extensions'', verb-suffixes used to modify the meaning of the verb. These qualities contribute to the morphological richness of these languages --- a stark contrast with European languages.

\begin{table*}[]
    \centering
    \resizebox{0.99\textwidth}{!}{\begin{tabular}{cccclccccc}
        \toprule[0.13em]
         \multirow{2}{*}{\textbf{Language}} & \multirow{2}{*}{\textbf{ISO Code}} &
         \multirow{2}{*}{\textbf{Lang. Family}}  & \multirow{2}{*}{\textbf{\# Noun Classes}}&\multirow{2}{*}{\textbf{Sources}} &\multicolumn{3}{c}{\bf \benchmark~(En$\rightarrow$XX)} &  \multicolumn{2}{c}{\bf Monolingual Data} \\
         \cmidrule(lr){6-8}\cmidrule(lr){9-10}
         &&&&& \textbf{Train} & \textbf{Valid} & \textbf{Test} & \textbf{Gold} & \textbf{Pseudo}\\
         \midrule[0.07em]
         Afrikaans & Af & Indo-European& ---  &J, O & 743K & 3000 & 3000 &1.3G & ---\\
         Bemba &  Bem  & Niger-Congo   & 9/6/15&  J & 275K & 3000 & 3000 & 38M & 1.0G \\
         Lingala & Ln  & Niger-Congo   & 9/6/15& J & 382K & 3000 & 3000 & 67M & 1.4G \\
         Rundi &  Run  & Niger-Congo   & 9/7/16&  J & 253K & 3000 & 3000 & 26M & 1.1G\\
         Sesotho & St    & Niger-Congo & 6/5/11&   J & 595K & 3000 & 3000 & 84M & 1.0G \\
         Swahili & Sw  & Niger-Congo   & 9/9/18& J, P & 700K & 3000 & 3000 &1.8G & 1.2G \\
         Xhosa &  Xh   & Niger-Congo   & 8/7/15&  J, X, M, Q & 610K& 3000 & 3000 & 203M & 1.2G\\
         Zulu &  Zu    & Niger-Congo   & 6/10/16&   J &664K & 3000 & 3000 & 121M & 1.4G \\
         \bottomrule[0.13em]
    \end{tabular}}
    \caption{Language characteristic and dataset statistics for \benchmark. Statistics for \benchmark~are measured in term of sentences. Monolingual data sizes are measured on the raw, pretokenized corpora. We abbreviate the sources for our benchmark as follows: J=JW300, O=OpenSubtitles, P=ParaCrawl, X=XhosaNavy, M=Memat, Q=QED. The \emph{\# Noun Classes} column shows the number of singular/plural/total noun classes.} \vspace{-0.2cm}
    \label{tb:data}
\end{table*}

\subsection{Data Sources} \label{sec:datasources}
For our benchmark, we leverage existing parallel data for each of our language pairs. This data is derived from two main sources: (1) open-source repository of parallel corpora, OPUS\footnote{\url{http://opus.nlpl.eu/}} \citep{tiedemann-2012-parallel} and (2) ParaCrawl \citep{espla-etal-2019-paracrawl}. From OPUS, we use the JW300 corpus \citep{agic-vulic-2019-jw300}, OpenSubtitles \citep{lison-tiedemann-2016-opensubtitles2016}, XhosaNavy, Memat, and QED \citep{abdelali-etal-2014-amara}. Despite the existence of this parallel data, these text datasets were often collected from large, relatively unclean multilingual corpora, e.g.~JW300 which was extracted from Jehovah's Witnesses text, or QED which was extracted from transcribed educational videos. This leads to many sentences with high lexical overlap, inconsistent tokenization, and other undesirable properties for a clean, reproducible benchmark.
\subsection{Data Preparation} \label{sec:preparation}
Training machine translation systems with small and noisy corpora for low-resource languages is challenging, and often leads to inaccurate translations. These noisy examples include sentences which contain only symbols and numbers, sentences which only consist of one token, sentences which are the same in both the source and target sides, etc. Furthermore, in these noisy extractions from large multilingual corpora such as JW300, there is a key issue of large text overlap over sentences. Given the risk of data leakage, this prevents one from naively splitting the corpus into random train/validation/test splits.  

To mitigate these issues, when preparing our data, we use a combination of automatic filtering techniques and manual human verification at each step to produce clean parallel data for the construction of our benchmark. For consistency across language pairs, we perform cleaning mainly based on the English side of the noisy parallel corpora. We list the automatic filtering techniques below: 
\paragraph{Removal of extremely short sentences} Since we focus on sentence-level machine translation,%
\footnote{While document-level translation is undoubtedly important, accuracy on the languages in \benchmark{} is still at the level where sentence-level translation is sufficiently challenging.}
we remove sentences containing less than three whitespace-tokenized tokens excluding numerical symbols and punctuation. Additionally, we remove pairs that contain no source or target sentences.

\paragraph{Removal of non-sentences} We remove sentences containing no letters, i.e., pairs that contain only numbers and symbols.
\paragraph{Tokenization normalization} We perform detokenization on all corpora using the detokenization script provided in the Moses \citep{koehn-etal-2007-moses} toolkit\footnote{\url{https://github.com/moses-smt/mosesdecoder/}}. Given that we collect data from various sources, this step is important to allow for consistent tokenization across corpora. 
\paragraph{Removal of sentences with high text overlap} To prevent data leakage, we remove sentences with high text overlap. To do this, we use Levenshtein-based fuzzy string matching\footnote{\url{https://github.com/maxbachmann/RapidFuzz}} and remove sentences that have a similarity score of over 60.
Given that measuring this score against all sentences in a corpus grows quadratically with respect to corpus length, we use the following two heuristics to remove sentences with high overlap in an efficient manner: (1) scoring similarity between the 50 alphabetically-sorted previous sentences,
(2): extracting the top 100K four-grams and performing the similarity score within each group of sentences containing at least one instance of a certain four-gram. 

\paragraph{Data Split} The resulting benchmark is constructed using the data that passes our automatic filtering checks, and we further split the data into \textit{train}, \textit{validation}, and \textit{test} for each language pair. We select 3,000 sentences with the least four-gram overlap (with the corpus) for both validation and testing while leaving the rest of the corpus to be used for training. Validation and test sentences are all further verified for quality. The resulting dataset statistics for each language pair can be seen in Table~\ref{tb:data}.

\subsection{Impact of Cleaning Process}
Given the non-trivial cleaning process and standardization of key components, such as tokenization/splits/data leakage, this cleaning provides a better representative corpus for the languages considered. We demonstrate this with an experiment comparing a randomly initialized English-Zulu models trained on (a) the original noisy data (including some test data leakage), (b) a model trained on noisy data (without data leakage) similar to the cleaning process used by \citet{nekoto-etal-2020-participatory}, and (c) a model trained on the AfroMT data. Scores for each setting are measured in BLEU on the clean test set: (a) 38.6, (b) 27.6, (c) 34.8.

Comparing the noisy model and the AfroMT model, we find that not filtering the data for leakage leads to misleading results, unreliably evaluating models on these LRLs. Additionally, as shown by (b) vs (c), not filtering for other artifacts hinders performance leading to unrealistically weak performance. Additional quantification of data leakage can be found in the Appendix.

\section{\afrobart} \label{sec:methods}
Given that we aim to provide strong baselines for our benchmark, we resort to multilingual sequence-to-sequence training. However, existing pretraining techniques have often been focused on the situation where monolingual data can be found in a larger quantity than parallel data. In this section we describe our proposed multilingual sequence-to-sequence pretraining techniques developed for the novel scenario where even monolingual data is scarce.

\subsection{Existing Methods}
\begin{figure}[th]
    \centering
    \includegraphics[width=\linewidth]{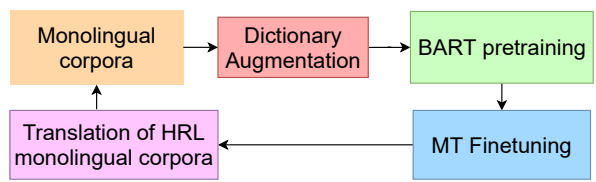}
    \caption{Iterative approach to pretraining using pseudo monolingual data and dictionaries}\vspace{-0.5cm}%
    \label{fig:method}
\end{figure}

The most widely used methods for multilingual sequence-to-sequence pretraining \citep{song2019mass,xue2020t5,liu2020mbart} make a core assumption that the amount of monolingual data in all languages exceeds the amount of parallel data. However, in the case of many African languages, digitized textual data is not widely available, leading this approach to be less effective in these scenarios as shown in Table~\ref{tb:results}. To mitigate this issue, we build on existing denoising pretraining techniques, particularly BART \citep{lewis-etal-2020-bart,liu2020mbart} and propose two data augmentation methods using dictionaries to augment high-resource monolingual data (\S\ref{sec:dict_augment}), and leveraging pseudo monolingual data in low-resource languages (\S\ref{sec:pseudomono}). Finally, we iterate the data augmentation with the model training (\S\ref{sec:iterative}) as shown in Figure~\ref{fig:method}.

\subsection{Dictionary Augmentation}
\label{sec:dict_augment}

Given that existing monolingual corpora in low-resource languages are small, we aim to increase the usage of words from the low-resource language in diverse contexts. To do so, we propose to take sentences from a high-resource language, and replace the words by their corresponding translations that are available in a dictionary extracted from our parallel corpora.
\paragraph{Dictionary Extraction} As our data augmentation technique requires a dictionary, we propose to extract the dictionary from parallel corpora using a statistical word aligner, \texttt{eflomal}\footnote{\url{https://github.com/robertostling/eflomal/}} \citep{Ostling2016efmaral}. Once we produce word alignments between tokens in our parallel corpora, we simply take word alignments that appear over 20 times to produce our bilingual dictionary.

\paragraph{Monolingual Data Augmentation} We assume to have access to three sources of data, i.e., high-resource corpus $H = \{H_0,\dots,H_T\}$, low-resource corpus $L = \{L_0,\dots,L_M\}$, and bilingual dictionary $D = \{(D_{0}^h, D_{0}^l),\dots,(D_{N_d}^h, D_{N_d}^l)\}$ with $N_d$ pairs mapping high-resource term $D_i^h$ to low-resource term $D_i^l$. Given this, for every high-resource sentence $H_i$ we replace 30\% of the tokens that match the high-resource terms contained in $D$ to their respective low-resource terms. In the case that there exists more than one low-resource term in $D_i^l$, we randomly select one to replace the high-resource term. Notably, with the assumption that high-resource monolingual data is more diverse in its content given its greater size, this augmentation technique is an effective method to increase the coverage of words from the low-resource lexicon in diverse settings.
\begin{figure}[t]
    \centering
    \includegraphics[width=\linewidth]{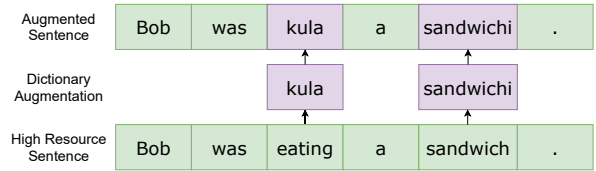}
    \caption{Transforming monolingual high-resource data to augmented code-switched data using an English-Swahili bilingual dictionary}
    \label{fig:dictionary_replace}
\end{figure}%

\subsection{Leveraging Pseudo-Monolingual Data} \label{sec:pseudomono}

Although leveraging dictionaries to produce code-switched monolingual data is a useful technique to introduce low-resource words 
in a wider variety of contexts, the code-switched sentences still lack the fluency and consistency of pure monolingual data. To further mitigate these fluency and data scarcity issues in the LRL, we propose to create fluent pseudo-monolingual data by translating the HRL monolingual data to the low-resource language using a pretrained machine translation model. %

Specifically, given a pretrained sequence-to-sequence model $\mM$, we finetune $\mM$ for the translation from HRL to LRL on a parallel corpus, i.e., $\gD_{ft} = \{(\gD_0^h,\gD_0^l),\dots,(\gD_{N_{ft}}^h,\gD_{N_{ft}}^l)\}$, and obtain a machine translation model $\mM_{ft}$. With the pretrained translation model $\mM_{ft}$, we then proceed to translate sentences from high-resource corpus $H$ to our low-resource language $l$ to produce pseudo LRL monolingual corpus $\tilde{L}$:
\begin{equation}
    \tilde{L} = \mM_{ft}(H; \Theta_{ft})
\end{equation}
Following this, we concatenate the existing low-resource corpus $L$ with $\tilde{L}$ and continue training our pretrained sequence-to-sequence model on this new pseudo-monolingual corpora.\footnote{Note that while this data generation process results in pseudo-parallel data, we do not experiment with using it in a supervised training scenario due to its noisy properties and in order to keep our benchmark comparable. However, this is an interesting direction which we leave for future work.}

\subsection{Iterative Multilingual Denoising Pretraining} \label{sec:iterative}
Given the pseudo-monolingual data synthesis step detailed in the previous \S\ref{sec:pseudomono}, we can simply transform this into an iterative pretraining procedure \citep{tran2020crosslingual}. That is, given the monolingual data synthesis procedure, we can leverage this procedure to produce a cycle in which a pretrained model is used to initialize an MT model to synthesize pseudo monolingual data and the produced data is used to further train the pretrained model (depicted in Figure~\ref{fig:method}). %

\section{Experimental Setup}
\label{sec:experiments}
In this section, we describe our experimental setup for both pretraining and finetuning strong baselines for our benchmark. Furthermore, we look to evaluate the efficacy of our proposed pretraining techniques and see whether they provide an impact on downstream performance on \benchmark.

\subsection{Pretraining}
\paragraph{Dataset} We pretrain \afrobart~on 11 languages: Afrikaans, English, French, Dutch\footnote{We select English and French due to their commonplace usage on the continent, as well as Dutch due to its similarity with Afrikaans.}, Bemba, Xhosa, Zulu, Rundi, Sesotho, Swahili, and Lingala. To construct the original monolingual corpora, we use a combination of the training sets in \benchmark~and data derived from CC100\footnote{\url{http://data.statmt.org/cc-100/}} \citep{wenzek-etal-2020-ccnet,conneau-etal-2020-unsupervised}. We only perform dictionary augmentation on our English monolingual data. We list monolingual and pseudo-monolingual corpora statistics in Table~\ref{tb:data}.

\paragraph{Balancing data across languages} As we are training on different languages with widely varying amounts of text, we use the exponential sampling technique used in \citet{lample2019crosslingual,liu2020mbart}, where the text is re-sampled according to smoothing parameter $\alpha$ as shown below:
\begin{equation}
 q_k = \frac{p_k^\alpha}{\sum_{j=1}^N p_j^\alpha}
\end{equation}
where $q_k$ refers to the re-sample probability for language $k$, given multinomial distribution $\{q_k\}_{k=1\ldots N}$ with original sampling probability $p_k$\footnote{$p_k$ is proportional to the amount of data for the language; in the case that we use dictionary augmented data, we keep $p_k$ proportional to the original data for the language}. As we work with many extremely low-resource languages, we choose smoothing parameter $\alpha=0.25$ (compared with the $\alpha=0.7$ used in mBART) to alleviate model bias towards an overwhelmingly higher proportion of data in the higher-resource languages.

\begin{table*}[]
    \centering
    \normalsize
    \scalebox{0.90}{\begin{tabular}{lcccccccccc}
    \toprule[0.15em]
     \textbf{Direction} & \multicolumn{2}{c}{\textbf{En-Run}} & \multicolumn{2}{c}{\textbf{En-Zu}} & \multicolumn{2}{c}{\textbf{En-Af}} & \multicolumn{2}{c}{\textbf{En-Xh}} \\
     &\textbf{BLEU} & \textbf{chrF} &\textbf{BLEU} & \textbf{chrF} &\textbf{BLEU} & \textbf{chrF} &\textbf{BLEU} & \textbf{chrF}\\
     \midrule[0.07em]
     Random& 22.92 & 51.89 & 34.84 & 65.54 & 48.33 & 68.11 & 24.36 & 52.91 \\
     mNMT & 21.53 & 50.62 & 31.53 & 62.95 & 43.39 & 64.73 & 22.28 & 54.81 \\
     \afrobartbaseline & 24.33 & 52.87 & \textbf{35.59} & 66.14  & 49.09 & 68.54 & 25.65 & 58.09\\
     \afrobartdict & 24.42 & 53.22  & 35.48 & 66.16 & 49.25 & 68.75 & 25.77 & 58.15 \\
     \afrobart & \textbf{24.62} & \textbf{53.24} & 35.58 & \textbf{66.30} & \textbf{49.80} & \textbf{69.03} & \textbf{25.80} & \textbf{58.22} \\
     \midrule[0.1em]
     \textbf{Direction}&\multicolumn{2}{c}{\textbf{En-Ln}} & \multicolumn{2}{c}{\textbf{En-Bem}} & \multicolumn{2}{c}{\textbf{En-St}} & \multicolumn{2}{c}{\textbf{En-Sw}} \\
     &\textbf{BLEU} & \textbf{chrF} &\textbf{BLEU} & \textbf{chrF} &\textbf{BLEU} & \textbf{chrF} &\textbf{BLEU} & \textbf{chrF}\\

     \midrule[0.07em]
     Random&    28.23 & 52.62 & 18.96 & 45.85 & 43.04 & 62.68 & 33.61 & 58.56 \\
     mNMT & 27.29 & 53.16 & 18.54 & 46.20 & 40.26 & 60.65 & 30.55 &  56.44 \\
     \afrobartbaseline & 29.12 & 54.31 & 20.07 & 47.50 & 43.79 & 63.22 & 34.19 & 59.08\\
     \afrobartdict   & 29.13 & 54.40 & 20.48 & 47.69 & 43.74 & 63.33 & 34.30 & 59.08 \\
     \afrobart &\textbf{29.46} & \textbf{54.68} &  \textbf{20.60} & \textbf{48.00} & \textbf{43.87} & \textbf{63.42} & \textbf{34.36} & \textbf{59.11} \\
    \bottomrule[0.15em]
    \end{tabular}}
    \caption{Results on \benchmark's En-XX Machine Translation}
    \label{tb:results}
\end{table*}

\paragraph{Hyperparameters} We use the following setup to train our AfroBART models, utilizing the mBART implementation in the \texttt{fairseq}\footnote{\url{https://github.com/pytorch/fairseq}} library \citep{ott-etal-2019-fairseq}. We tokenize data using SentencePiece \citep{Kudo_2018}, using a 80K subword vocabulary. We use the Transformer-base architecture of a hidden dimension of 512, feedforward size of 2048, and 6 layers for both the encoder and decoder. We set the maximum sequence length to be 512, using a batch size of 1024 for 100K iterations with 32 NVIDIA V100 GPUs for one day. When we continue training using pseudo-monolingual data, we use a learning rate of $7\times10^{-5}$ and warm up over 5K iterations and train for 35K iterations.

\subsection{Finetuning}
\paragraph{Baselines} We use the following baselines for our benchmark:
\begin{itemize}
    \item \textbf{\afrobartbaseline} We pretrain a model using only the original monolingual corpora in a similar fashion to \citet{liu2020mbart}.\vspace{-0.12cm} %
    \item \textbf{\afrobartdict} We pretrain a model using the original data in addition to a dictionary augmented English monolingual corpora in Afrikaans, Bemba, Sesotho, Xhosa, Zulu, Lingala, and Swahili.\vspace{-0.12cm}
    \item \textbf{\afrobart} We continue training the dictionary augmented AfroBART model, using pseudo monolingual data produce by its finetuned counterparts. Due to computational constraints we only perform one iteration of our iterative approach. Statistics for the pseudo-monolingual data can be seen in Table~\ref{tb:data}.\vspace{-0.12cm}
    \item \textbf{Cross-Lingual Transfer (CLT)} When experimenting on the effect of pretraining with various amounts of finetuning data, we use strong cross-lingual transfer models, involving training from scratch on a combination of both our low-resource data and a similar relatively high-resource language following \citet{neubig-hu-2018-rapid}. \vspace{-0.12cm}
    \item \textbf{Multilingual Neural Machine Translation (mNMT)} We also experiment with a vanilla multilingual machine translation system \citep{dabre2020multilingual} trained on all En-XX directions.  \vspace{-0.12cm}
    \item \textbf{Random} As additional baselines, we also provide a comparison with a randomly initialized Transformer-base \citep{vaswani2017attention} models for each translation pair.\vspace{-0.12cm}
\end{itemize}

\paragraph{Evaluation} We evaluate our system outputs using two automatic evaluation metrics: detokenized BLEU \citep{papineni-etal-2002-bleu,post-2018-call} and chrF \citep{popovic-2015-chrf}. Although BLEU is a standard metric for machine translation, being cognizant of the morphological richness of the languages in the \benchmark~benchmark, we use chrF to measure performance at a character level. Both metrics are measured using the SacreBLEU library\footnote{\url{https://github.com/mjpost/sacrebleu}} \citep{post-2018-call}. 
\begin{figure*}[t]
    \centering
    \includegraphics[width=0.8\linewidth]{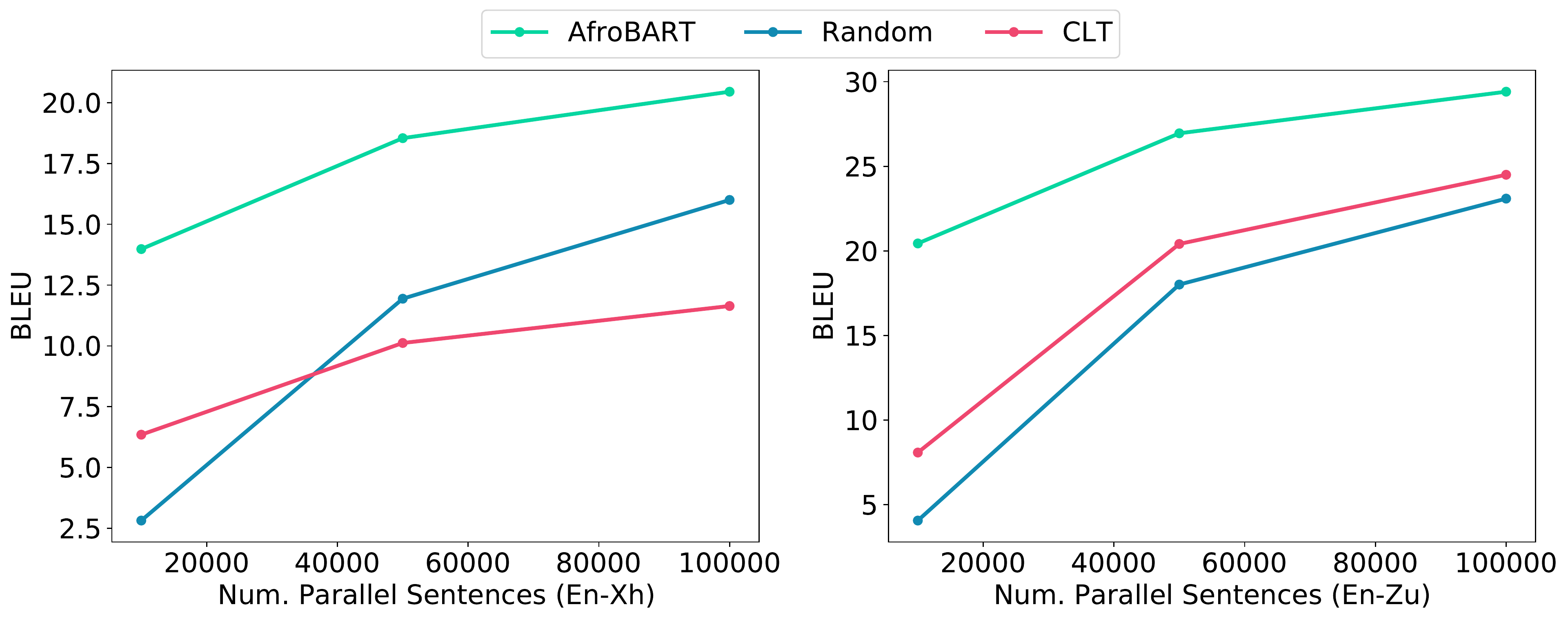}
    \caption{Visualization of results using various amounts of parallel data on English-Xhosa and English-Zulu. We compare \afrobart, random initialization and cross-lingual transfer.} \vspace{-0.2cm} %
    \label{fig:limited_data}
\end{figure*}
\section{Results and Discussion}
\subsection{Performance on En-XX Translation}
Table~\ref{tb:results} shows the results on En-XX translation on the \benchmark~benchmark comparing random initialization with various pretrained AfroBART configurations. We find that initializing with pretrained AfroBART weights results in performance gains of $\sim$1 BLEU across all language pairs. Furthermore, we observe that augmenting our pretraining data with a dictionary results in performance gains across all pairs in terms of chrF and 6/8 pairs in terms of BLEU. The gain is especially clear on languages with fewer amounts of monolingual data such as Rundi and Bemba, demonstrating the effectiveness of our data augmentation techniques on low-resource translation. Moreover we see further improvements when augmenting with pseudo monolingual data, especially on pairs with fewer data which validates the usage of this technique.
\subsection{Performance vs Amount of Parallel Data}
We perform experiments to demonstrate the effect on pretraining with various amounts of parallel data (10k, 50k, and 100k pairs) on two related language pairs: English-Xhosa and English-Zulu. We compare AfroBART (with both dictionary augmentation and pseudo monolingual data) with randomly initialized models, and cross-lingual transfer models \citep{neubig-hu-2018-rapid} jointly trained with a larger amount of parallel data (full \benchmark~data) in a related language.

In Figure~\ref{fig:limited_data}, a pretrained \afrobart~model finetuned on 10K pairs can almost double the performance of other models (with a significant performance increase over random initialization of 15+ BLEU on English-Zulu), outperforming both cross-lingual transfer and randomly initialized models trained on 5x the data. Furthermore, we notice that CLT performs than Random on English-Xhosa as the data size increases. Although we do not have an exact explanation for this, we believe this has to do with the other language data adding noise rather than additional supervision as the data size increases. We detail these results in Table 3 of the Appendix. %

\paragraph{Comparison on convergence speed} In contrast to the cross-lingual transfer baseline which involves the usage of more data, and the random initialization baseline which needs to learn from scratch, AfroBART is able to leverage the knowledge gained during training for fast adaptation even with small amounts of data. For example, AfroBART converged within 1,000 iterations when finetuning on 10K pairs on English-Zulu, whereas the random initialization and cross-lingual transfer baselines converged within 2.5K and 12K iterations respectively. This is promising as it indicates that we can leverage these models quickly for other tasks where there is much fewer parallel data.

\subsection{Fine-grained Language Analysis}
We further provide a suite of fine-grained analysis tools to compare the baseline systems. In particular, we are interested in evaluating the translation accuracy of noun classes in the considered African languages in the Niger-Congo family, as these languages are morphologically rich and often have more than 10 classes based on the prefix of the word. For example, \textit{kitabu}  and \textit{vitabu} in Swahili refer to \textit{book} and \textit{books} in English, respectively. Based on this language characteristic, our fine-grained analysis tool calculates the translation accuracy of the nouns with the top 10 most frequent prefixes in the test data. To do so, one of the challenges is to identify nouns in a sentence written in the target African language. However, there is no available part-of-speech (POS) tagger for these languages. To tackle this challenge, we propose to use a label projection method based on word alignment. Specifically, we first leverage an existing English POS tagger in the \texttt{spaCy}\footnote{\url{https://spacy.io/}} library to annotate the English source sentences. We then use the \texttt{fast\_align}\footnote{\url{https://github.com/clab/fast_align}} tool~\citep{dyer-etal-2013-simple} to train a word alignment model on the training data for the En-XX language pair, and use the alignment model to obtain the word-level alignment for the test data. We assign the POS tags of the source words in English to their aligned target words in the African language. We then measure the translation accuracy of the nouns in the African language by checking whether the correct nouns are included in the translated sentences by systems in comparison. Notably, our analysis tool can also measure the translation accuracy of the words in the other POS tags, (e.g.~verbs, adjectives) which are often adjusted with different noun classes.

\begin{figure}[t]
    \centering
    \includegraphics[width=0.8\linewidth]{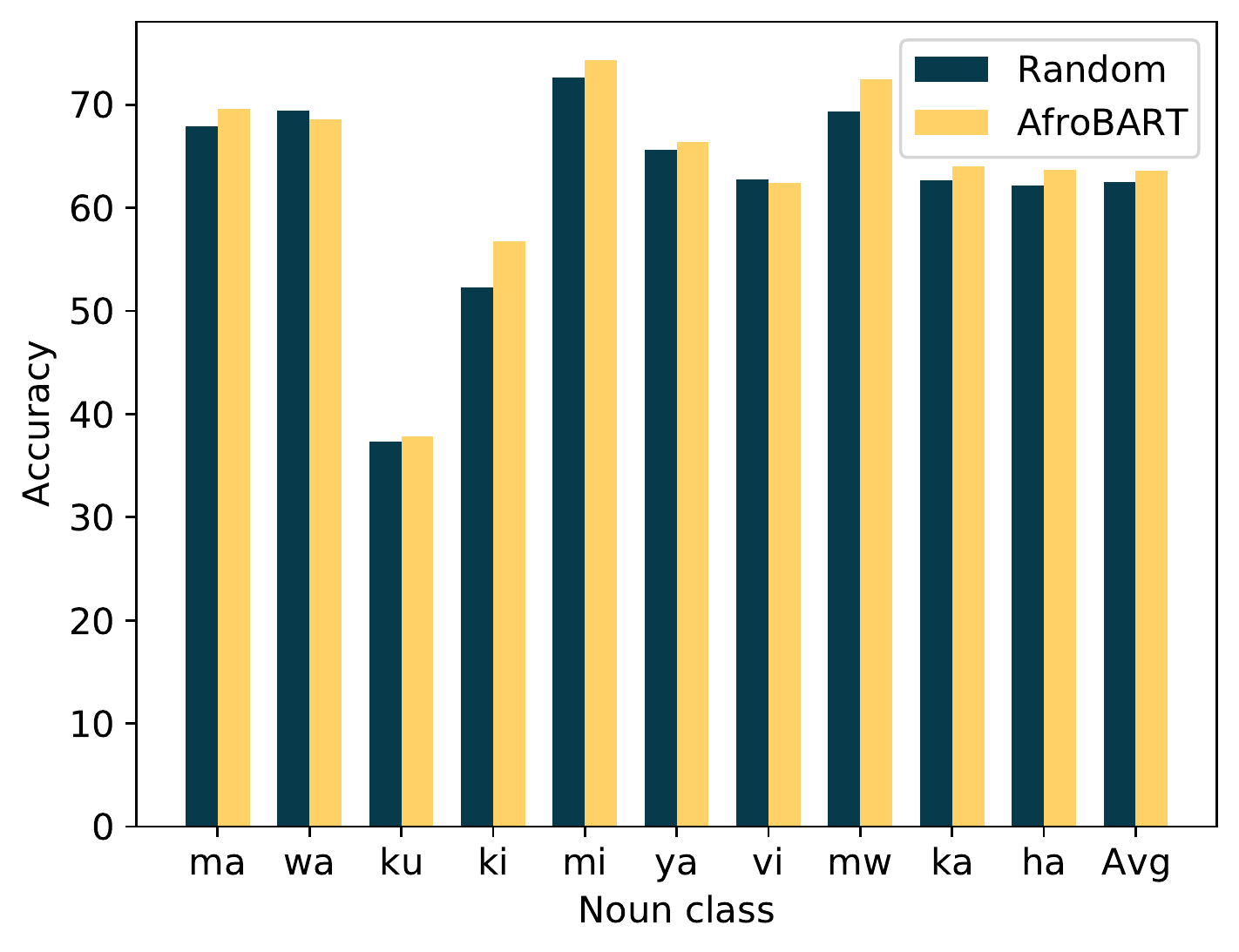}
    \caption{Translation accuracy of the AfroBART and Random baseline systems on Swahili noun classes with top 10 most frequent 2-character prefixes. } 
    \label{fig:swahili_acc}
\end{figure}

Figure~\ref{fig:swahili_acc} compares the AfroBART and Random baseline in terms of translation accuracy of nouns in Swahili. First, we find that both systems perform worse on translating nouns with the prefix ``ku-'' which usually represent the infinitive form of verbs, e.g., \textit{kula} for \textit{eating}. Secondly, we find that AfroBART significantly improves translation accuracy for nouns with prefixes ``ki-'' (describing man-made tools/languages, e.g., \textit{kitabu} for \textit{book}) and ``mw-'' (describing a person, e.g., \textit{mwalimu} for \textit{teacher}). Finally, AfroBART improves the translation accuracy on average over the ten noun classes by 1.08\% over the Random baseline.

\begin{figure}[t]
    \centering
    \includegraphics[width=0.8\linewidth]{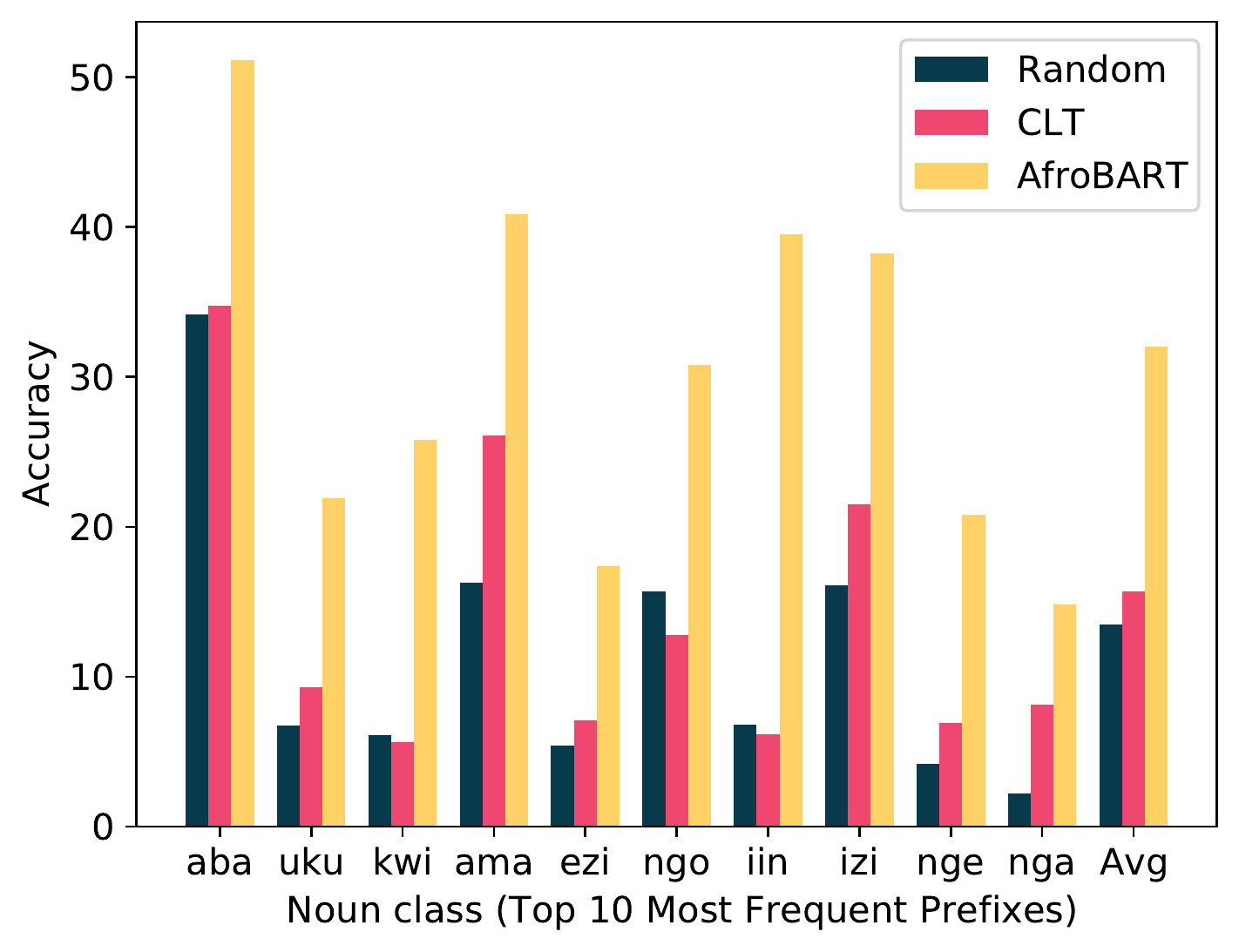}
    \caption{Translation accuracy of the AfroBART and Random baseline systems on Xhosa (10k pairs) noun classes with top 10 most frequent 3-character prefixes.}
    \label{fig:xhosa_acc}
\end{figure}

We also perform this analysis on our data-constrained scenario for English-Xhosa, shown in Figure~\ref{fig:xhosa_acc}. It can be seen that leveraging cross-lingual transfer (trained on both Xhosa and Zulu) models improved noun class accuracy on classes such as \textit{uku} (infinitive noun class), \textit{izi} (plural for objects), and \textit{ama} (plural for body parts) which are shared between languages. This can be contrasted with \textit{iin} (plural for animals) which is only used in Xhosa, where CLT decreases performance. These analyses which require knowledge of unique grammar found in these langauges can be used for diagnosing cross-lingual transfer for these langauges. 
Also, we note that AfroBART almost doubles the accuracy (improvement of 16.33\%) of the cross-lingual transfer baseline on these noun classes.

\subsection{Shortcomings of \benchmark}
Although we believe \benchmark~to be an important step in the right direction, we acknowledge it is far from being the end-all-be-all. Specifically, we note the following: (1) the lack of domain diversity among many languages (being largely from religious oriented corpora) and (2) the corpora may still contain some more fine-grained forms of noise in terms of translation given its origin. Given this, in the future we look to include more diverse data sources and more languages and encourage the community to do so as well.

\section{Related Work}
\paragraph{Machine Translation Benchmarks} Previous work in benchmarking includes the commonly used WMT \citep{barrault-etal-2020-findings} and IWSLT \citep{federico-etal-2020-speech} shared tasks.  Recent work on MT benchmarks for low-resource languages, such as that of \citet{guzman2019he}, have been used for the purpose of studying current NMT techniques for low-resource languages. 
\paragraph{Multilingual Pretraining} Multilingual encoder pretraining \citep{devlin-etal-2019-bert,lample2019crosslingual,conneau-etal-2020-unsupervised} has been demonstrated to be an effective technique for cross-lingual transfer on a variety of classification tasks \citep{pmlr-v119-hu20b,artetxe2020crosslingual}. More recently, sequence-to-sequence pretraining has emerged as a prevalent method for achieving better performance \citep{lewis-etal-2020-bart,song2019mass} on generation tasks.
\citet{liu2020mbart} proposed a multilingual approach to BART \citep{lewis-etal-2020-bart} and demonstrated increased performance on MT. Building on these works, we extend this to a LRL-focused setting, developing two new techniques for improved performance given monolingual data-scarcity. In concurrent work, \citet{liu2021continual,reid2021paradise} also look at using code-switched corpora for sequence-to-sequence pre-training.
\paragraph{NLP for African Languages} Benchmarking machine translation for African languages was first done by \citet{abbott-martinus-2019-benchmarking} for southern African languages and \citet{abate-etal-2018-parallel} for Ethiopian languages. Recent work in NLP for African languages has largely revolved around the grassroots translation initiative Masakhane \citep{orife2020masakhane,nekoto-etal-2020-participatory}. This bottom-up approach to dataset creation \citep{nekoto-etal-2020-participatory}, while very valuable, has tended to result in datasets with somewhat disparate data splits and quality standards. 
In contrast, \benchmark~provides a cleaner corpus for the 8 supported languages. We plan to open source the the entire benchmark (splits included) to promote reproducible results in the community. 
\section{Conclusion}
In this work we proposed a standardized, clean, and reproducible benchmark for 8 African languages, \benchmark, as well as
novel pretraining strategies in the previously unexplored low-resource focused setting. Our benchmark and evaluation suite are a step towards larger, reproducible benchmarks in these languages, helping to provide insights on how current MT techniques work for these under-explored languages. We will release this benchmark, our pretrained AfroBART models, dictionaries, and pseudo monolingual data to the community to facilitate further work in this area.

In future work we look to use similar methodology to advance in both of these directions. We look to increase the number of language pairs in \benchmark~to be more representative of the African continent. Additionally, we look to scale up our pretraining approaches for increased performance.
\section*{Acknowlegements}
We thank Antonios Anastasopoulos and Edison Marrese-Taylor, and the anonymous reviewers for feedback and comments. We also thank Aditi Chaudhary and Kathleen Siminyu for helpful discussions in early stages of this work. MR is grateful to the Masason Foundation for their support.
\bibliographystyle{acl_natbib}
\bibliography{acl2021}

\begin{thebibliography}{46}
\expandafter\ifx\csname natexlab\endcsname\relax\def\natexlab#1{#1}\fi

\bibitem[{Abate et~al.(2018)Abate, Melese, Tachbelie, Meshesha, Atinafu,
  Mulugeta, Assabie, Abera, Ephrem, Abebe, Tsegaye, Lemma, Andargie, and
  Shifaw}]{abate-etal-2018-parallel}
Solomon~Teferra Abate, Michael Melese, Martha~Yifiru Tachbelie, Million
  Meshesha, Solomon Atinafu, Wondwossen Mulugeta, Yaregal Assabie, Hafte Abera,
  Binyam Ephrem, Tewodros Abebe, Wondimagegnhue Tsegaye, Amanuel Lemma, Tsegaye
  Andargie, and Seifedin Shifaw. 2018.
\newblock \href {https://aclanthology.org/C18-1262} {Parallel corpora for
  bi-lingual {E}nglish-{E}thiopian languages statistical machine translation}.
\newblock In \emph{Proceedings of the 27th International Conference on
  Computational Linguistics}, pages 3102--3111, Santa Fe, New Mexico, USA.
  Association for Computational Linguistics.

\bibitem[{Abbott and Martinus(2019)}]{abbott-martinus-2019-benchmarking}
Jade Abbott and Laura Martinus. 2019.
\newblock \href {https://aclanthology.org/W19-3632} {Benchmarking neural
  machine translation for {S}outhern {A}frican languages}.
\newblock In \emph{Proceedings of the 2019 Workshop on Widening NLP}, pages
  98--101, Florence, Italy. Association for Computational Linguistics.

\bibitem[{Abdelali et~al.(2014)Abdelali, Guzman, Sajjad, and
  Vogel}]{abdelali-etal-2014-amara}
Ahmed Abdelali, Francisco Guzman, Hassan Sajjad, and Stephan Vogel. 2014.
\newblock \href
  {http://www.lrec-conf.org/proceedings/lrec2014/pdf/877_Paper.pdf} {The
  {AMARA} corpus: Building parallel language resources for the educational
  domain}.
\newblock In \emph{Proceedings of the Ninth International Conference on
  Language Resources and Evaluation ({LREC}'14)}, pages 1856--1862, Reykjavik,
  Iceland. European Language Resources Association (ELRA).

\bibitem[{Agi{\'c} and Vuli{\'c}(2019)}]{agic-vulic-2019-jw300}
{\v{Z}}eljko Agi{\'c} and Ivan Vuli{\'c}. 2019.
\newblock \href {https://doi.org/10.18653/v1/P19-1310} {{JW}300: A
  wide-coverage parallel corpus for low-resource languages}.
\newblock In \emph{Proceedings of the 57th Annual Meeting of the Association
  for Computational Linguistics}, pages 3204--3210, Florence, Italy.
  Association for Computational Linguistics.

\bibitem[{Arivazhagan et~al.(2019)Arivazhagan, Bapna, Firat, Lepikhin, Johnson,
  Krikun, Chen, Cao, Foster, Cherry et~al.}]{arivazhagan2019massively}
Naveen Arivazhagan, Ankur Bapna, Orhan Firat, Dmitry Lepikhin, Melvin Johnson,
  Maxim Krikun, Mia~Xu Chen, Yuan Cao, George Foster, Colin Cherry, et~al.
  2019.
\newblock Massively multilingual neural machine translation in the wild:
  Findings and challenges.
\newblock \emph{arXiv preprint arXiv:1907.05019}.

\bibitem[{Artetxe et~al.(2020)Artetxe, Ruder, and
  Yogatama}]{artetxe2020crosslingual}
Mikel Artetxe, Sebastian Ruder, and Dani Yogatama. 2020.
\newblock \href {https://doi.org/10.18653/v1/2020.acl-main.421} {On the
  cross-lingual transferability of monolingual representations}.
\newblock In \emph{Proceedings of the 58th Annual Meeting of the Association
  for Computational Linguistics}, pages 4623--4637, Online. Association for
  Computational Linguistics.

\bibitem[{Bahdanau et~al.(2015)Bahdanau, Cho, and Bengio}]{bahdanau2014eural}
Dzmitry Bahdanau, Kyunghyun Cho, and Yoshua Bengio. 2015.
\newblock \href {http://arxiv.org/abs/1409.0473} {Neural machine translation by
  jointly learning to align and translate}.
\newblock In \emph{3rd International Conference on Learning Representations,
  {ICLR} 2015, San Diego, CA, USA, May 7-9, 2015, Conference Track
  Proceedings}.

\bibitem[{Bendor-Samuel and Hartell(1989)}]{bendorsamuel1989niger}
John~T. Bendor-Samuel and Rhonda~L. Hartell, editors. 1989.
\newblock \emph{The Niger-Congo Languages: A classification and description of
  Africa's largest language family}. University Press of America, Lanham, MD.

\bibitem[{Bojar et~al.(2017)Bojar, Chatterjee, Federmann, Graham, Haddow,
  Huang, Huck, Koehn, Liu, Logacheva, Monz, Negri, Post, Rubino, Specia, and
  Turchi}]{barrault-etal-2020-findings}
Ond{\v{r}}ej Bojar, Rajen Chatterjee, Christian Federmann, Yvette Graham, Barry
  Haddow, Shujian Huang, Matthias Huck, Philipp Koehn, Qun Liu, Varvara
  Logacheva, Christof Monz, Matteo Negri, Matt Post, Raphael Rubino, Lucia
  Specia, and Marco Turchi. 2017.
\newblock \href {https://doi.org/10.18653/v1/W17-4717} {Findings of the 2017
  conference on machine translation ({WMT}17)}.
\newblock In \emph{Proceedings of the Second Conference on Machine
  Translation}, pages 169--214, Copenhagen, Denmark. Association for
  Computational Linguistics.

\bibitem[{Conneau et~al.(2020)Conneau, Khandelwal, Goyal, Chaudhary, Wenzek,
  Guzm{\'a}n, Grave, Ott, Zettlemoyer, and
  Stoyanov}]{conneau-etal-2020-unsupervised}
Alexis Conneau, Kartikay Khandelwal, Naman Goyal, Vishrav Chaudhary, Guillaume
  Wenzek, Francisco Guzm{\'a}n, Edouard Grave, Myle Ott, Luke Zettlemoyer, and
  Veselin Stoyanov. 2020.
\newblock \href {https://doi.org/10.18653/v1/2020.acl-main.747} {Unsupervised
  cross-lingual representation learning at scale}.
\newblock In \emph{Proceedings of the 58th Annual Meeting of the Association
  for Computational Linguistics}, pages 8440--8451, Online. Association for
  Computational Linguistics.

\bibitem[{Conneau and Lample(2019)}]{lample2019crosslingual}
Alexis Conneau and Guillaume Lample. 2019.
\newblock \href
  {https://proceedings.neurips.cc/paper/2019/hash/c04c19c2c2474dbf5f7ac4372c5b9af1-Abstract.html}
  {Cross-lingual language model pretraining}.
\newblock In \emph{Advances in Neural Information Processing Systems 32: Annual
  Conference on Neural Information Processing Systems 2019, NeurIPS 2019,
  December 8-14, 2019, Vancouver, BC, Canada}, pages 7057--7067.

\bibitem[{Dabre et~al.(2020)Dabre, Chu, and
  Kunchukuttan}]{dabre2020multilingual}
Raj Dabre, Chenhui Chu, and Anoop Kunchukuttan. 2020.
\newblock \href {https://doi.org/10.1145/3406095} {A survey of multilingual
  neural machine translation}.
\newblock \emph{ACM Comput. Surv.}, 53(5).

\bibitem[{Devlin et~al.(2019)Devlin, Chang, Lee, and
  Toutanova}]{devlin-etal-2019-bert}
Jacob Devlin, Ming-Wei Chang, Kenton Lee, and Kristina Toutanova. 2019.
\newblock \href {https://doi.org/10.18653/v1/N19-1423} {{BERT}: Pre-training of
  deep bidirectional transformers for language understanding}.
\newblock In \emph{Proceedings of the 2019 Conference of the North {A}merican
  Chapter of the Association for Computational Linguistics: Human Language
  Technologies, Volume 1 (Long and Short Papers)}, pages 4171--4186,
  Minneapolis, Minnesota. Association for Computational Linguistics.

\bibitem[{Dyer et~al.(2013)Dyer, Chahuneau, and Smith}]{dyer-etal-2013-simple}
Chris Dyer, Victor Chahuneau, and Noah~A. Smith. 2013.
\newblock \href {https://www.aclweb.org/anthology/N13-1073} {A simple, fast,
  and effective reparameterization of {IBM} model 2}.
\newblock In \emph{Proceedings of the 2013 Conference of the North {A}merican
  Chapter of the Association for Computational Linguistics: Human Language
  Technologies}, pages 644--648, Atlanta, Georgia. Association for
  Computational Linguistics.

\bibitem[{Espl{\`a} et~al.(2019)Espl{\`a}, Forcada, Ram{\'\i}rez-S{\'a}nchez,
  and Hoang}]{espla-etal-2019-paracrawl}
Miquel Espl{\`a}, Mikel Forcada, Gema Ram{\'\i}rez-S{\'a}nchez, and Hieu Hoang.
  2019.
\newblock \href {https://www.aclweb.org/anthology/W19-6721} {{P}ara{C}rawl:
  Web-scale parallel corpora for the languages of the {EU}}.
\newblock In \emph{Proceedings of Machine Translation Summit XVII Volume 2:
  Translator, Project and User Tracks}, pages 118--119, Dublin, Ireland.
  European Association for Machine Translation.

\bibitem[{Federico et~al.(2020)Federico, Enyedi, Barra-Chicote, Giri, Isik,
  Krishnaswamy, and Sawaf}]{federico-etal-2020-speech}
Marcello Federico, Robert Enyedi, Roberto Barra-Chicote, Ritwik Giri, Umut
  Isik, Arvindh Krishnaswamy, and Hassan Sawaf. 2020.
\newblock \href {https://doi.org/10.18653/v1/2020.iwslt-1.31} {From
  speech-to-speech translation to automatic dubbing}.
\newblock In \emph{Proceedings of the 17th International Conference on Spoken
  Language Translation}, pages 257--264, Online. Association for Computational
  Linguistics.

\bibitem[{Gordon et~al.(2021)Gordon, Duh, and Kaplan}]{gordon2021data}
Mitchell~A Gordon, Kevin Duh, and Jared Kaplan. 2021.
\newblock \href {https://openreview.net/forum?id=IKA7MLxsLSu} {Data and
  parameter scaling laws for neural machine translation}.
\newblock In \emph{ACL Rolling Review - May 2021}.

\bibitem[{Guzm{\'a}n et~al.(2019)Guzm{\'a}n, Chen, Ott, Pino, Lample, Koehn,
  Chaudhary, and Ranzato}]{guzman2019he}
Francisco Guzm{\'a}n, Peng-Jen Chen, Myle Ott, Juan Pino, Guillaume Lample,
  Philipp Koehn, Vishrav Chaudhary, and Marc{'}Aurelio Ranzato. 2019.
\newblock \href {https://doi.org/10.18653/v1/D19-1632} {The {FLORES} evaluation
  datasets for low-resource machine translation: {N}epali{--}{E}nglish and
  {S}inhala{--}{E}nglish}.
\newblock In \emph{Proceedings of the 2019 Conference on Empirical Methods in
  Natural Language Processing and the 9th International Joint Conference on
  Natural Language Processing (EMNLP-IJCNLP)}, pages 6098--6111, Hong Kong,
  China. Association for Computational Linguistics.

\bibitem[{Hu et~al.(2020)Hu, Ruder, Siddhant, Neubig, Firat, and
  Johnson}]{pmlr-v119-hu20b}
Junjie Hu, Sebastian Ruder, Aditya Siddhant, Graham Neubig, Orhan Firat, and
  Melvin Johnson. 2020.
\newblock \href {https://proceedings.mlr.press/v119/hu20b.html} {{XTREME}: A
  massively multilingual multi-task benchmark for evaluating cross-lingual
  generalisation}.
\newblock In \emph{Proceedings of the 37th International Conference on Machine
  Learning}, volume 119 of \emph{Proceedings of Machine Learning Research},
  pages 4411--4421. PMLR.

\bibitem[{Kingma and Ba(2015)}]{kingma2017adam}
Diederik~P. Kingma and Jimmy Ba. 2015.
\newblock \href {http://arxiv.org/abs/1412.6980} {Adam: {A} method for
  stochastic optimization}.
\newblock In \emph{3rd International Conference on Learning Representations,
  {ICLR} 2015, San Diego, CA, USA, May 7-9, 2015, Conference Track
  Proceedings}.

\bibitem[{Koehn et~al.(2007)Koehn, Hoang, Birch, Callison-Burch, Federico,
  Bertoldi, Cowan, Shen, Moran, Zens, Dyer, Bojar, Constantin, and
  Herbst}]{koehn-etal-2007-moses}
Philipp Koehn, Hieu Hoang, Alexandra Birch, Chris Callison-Burch, Marcello
  Federico, Nicola Bertoldi, Brooke Cowan, Wade Shen, Christine Moran, Richard
  Zens, Chris Dyer, Ond{\v{r}}ej Bojar, Alexandra Constantin, and Evan Herbst.
  2007.
\newblock \href {https://www.aclweb.org/anthology/P07-2045} {{M}oses: Open
  source toolkit for statistical machine translation}.
\newblock In \emph{Proceedings of the 45th Annual Meeting of the Association
  for Computational Linguistics Companion Volume Proceedings of the Demo and
  Poster Sessions}, pages 177--180, Prague, Czech Republic. Association for
  Computational Linguistics.

\bibitem[{Koehn and Knowles(2017)}]{koehn-knowles-2017-six}
Philipp Koehn and Rebecca Knowles. 2017.
\newblock \href {https://doi.org/10.18653/v1/W17-3204} {Six challenges for
  neural machine translation}.
\newblock In \emph{Proceedings of the First Workshop on Neural Machine
  Translation}, pages 28--39, Vancouver. Association for Computational
  Linguistics.

\bibitem[{Kudo and Richardson(2018)}]{Kudo_2018}
Taku Kudo and John Richardson. 2018.
\newblock \href {https://doi.org/10.18653/v1/D18-2012} {{S}entence{P}iece: A
  simple and language independent subword tokenizer and detokenizer for neural
  text processing}.
\newblock In \emph{Proceedings of the 2018 Conference on Empirical Methods in
  Natural Language Processing: System Demonstrations}, pages 66--71, Brussels,
  Belgium. Association for Computational Linguistics.

\bibitem[{Lewis et~al.(2020)Lewis, Liu, Goyal, Ghazvininejad, Mohamed, Levy,
  Stoyanov, and Zettlemoyer}]{lewis-etal-2020-bart}
Mike Lewis, Yinhan Liu, Naman Goyal, Marjan Ghazvininejad, Abdelrahman Mohamed,
  Omer Levy, Veselin Stoyanov, and Luke Zettlemoyer. 2020.
\newblock \href {https://doi.org/10.18653/v1/2020.acl-main.703} {{BART}:
  Denoising sequence-to-sequence pre-training for natural language generation,
  translation, and comprehension}.
\newblock In \emph{Proceedings of the 58th Annual Meeting of the Association
  for Computational Linguistics}, pages 7871--7880, Online. Association for
  Computational Linguistics.

\bibitem[{Lison and Tiedemann(2016)}]{lison-tiedemann-2016-opensubtitles2016}
Pierre Lison and J{\"o}rg Tiedemann. 2016.
\newblock \href {https://www.aclweb.org/anthology/L16-1147}
  {{O}pen{S}ubtitles2016: Extracting large parallel corpora from movie and {TV}
  subtitles}.
\newblock In \emph{Proceedings of the Tenth International Conference on
  Language Resources and Evaluation ({LREC}'16)}, pages 923--929,
  Portoro{\v{z}}, Slovenia. European Language Resources Association (ELRA).

\bibitem[{Liu et~al.(2020)Liu, Gu, Goyal, Li, Edunov, Ghazvininejad, Lewis, and
  Zettlemoyer}]{liu2020mbart}
Yinhan Liu, Jiatao Gu, Naman Goyal, Xian Li, Sergey Edunov, Marjan
  Ghazvininejad, Mike Lewis, and Luke Zettlemoyer. 2020.
\newblock \href {https://transacl.org/ojs/index.php/tacl/article/view/2107}
  {Multilingual denoising pre-training for neural machine translation}.
\newblock \emph{Transactions of the Association for Computational Linguistics},
  8(0):726--742.

\bibitem[{Liu et~al.(2021)Liu, Winata, and Fung}]{liu2021continual}
Zihan Liu, Genta~Indra Winata, and Pascale Fung. 2021.
\newblock \href {https://doi.org/10.18653/v1/2021.findings-acl.239} {Continual
  mixed-language pre-training for extremely low-resource neural machine
  translation}.
\newblock \emph{Findings of the Association for Computational Linguistics:
  ACL-IJCNLP 2021}.

\bibitem[{Meng et~al.(2019)Meng, Ren, Sun, Li, Yuan, Wu, and
  Li}]{meng2019large}
Yuxian Meng, Xiangyuan Ren, Zijun Sun, Xiaoya Li, Arianna Yuan, Fei Wu, and
  Jiwei Li. 2019.
\newblock Large-scale pretraining for neural machine translation with tens of
  billions of sentence pairs.
\newblock \emph{arXiv preprint arXiv:1909.11861}.

\bibitem[{Nekoto et~al.(2020)Nekoto, Marivate, Matsila, Fasubaa, Fagbohungbe,
  Akinola, Muhammad, Kabongo~Kabenamualu, Osei, Sackey, Niyongabo, Macharm,
  Ogayo, Ahia, Berhe, Adeyemi, Mokgesi-Selinga, Okegbemi, Martinus, Tajudeen,
  Degila, Ogueji, Siminyu, Kreutzer, Webster, Ali, Abbott, Orife, Ezeani,
  Dangana, Kamper, Elsahar, Duru, Kioko, Espoir, van Biljon, Whitenack,
  Onyefuluchi, Emezue, Dossou, Sibanda, Bassey, Olabiyi, Ramkilowan, {\"O}ktem,
  Akinfaderin, and Bashir}]{nekoto-etal-2020-participatory}
Wilhelmina Nekoto, Vukosi Marivate, Tshinondiwa Matsila, Timi Fasubaa, Taiwo
  Fagbohungbe, Solomon~Oluwole Akinola, Shamsuddeen Muhammad, Salomon
  Kabongo~Kabenamualu, Salomey Osei, Freshia Sackey, Rubungo~Andre Niyongabo,
  Ricky Macharm, Perez Ogayo, Orevaoghene Ahia, Musie~Meressa Berhe, Mofetoluwa
  Adeyemi, Masabata Mokgesi-Selinga, Lawrence Okegbemi, Laura Martinus,
  Kolawole Tajudeen, Kevin Degila, Kelechi Ogueji, Kathleen Siminyu, Julia
  Kreutzer, Jason Webster, Jamiil~Toure Ali, Jade Abbott, Iroro Orife, Ignatius
  Ezeani, Idris~Abdulkadir Dangana, Herman Kamper, Hady Elsahar, Goodness Duru,
  Ghollah Kioko, Murhabazi Espoir, Elan van Biljon, Daniel Whitenack,
  Christopher Onyefuluchi, Chris~Chinenye Emezue, Bonaventure F.~P. Dossou,
  Blessing Sibanda, Blessing Bassey, Ayodele Olabiyi, Arshath Ramkilowan, Alp
  {\"O}ktem, Adewale Akinfaderin, and Abdallah Bashir. 2020.
\newblock \href {https://doi.org/10.18653/v1/2020.findings-emnlp.195}
  {Participatory research for low-resourced machine translation: A case study
  in {A}frican languages}.
\newblock In \emph{Findings of the Association for Computational Linguistics:
  EMNLP 2020}, pages 2144--2160, Online. Association for Computational
  Linguistics.

\bibitem[{Neubig and Hu(2018)}]{neubig-hu-2018-rapid}
Graham Neubig and Junjie Hu. 2018.
\newblock \href {https://doi.org/10.18653/v1/D18-1103} {Rapid adaptation of
  neural machine translation to new languages}.
\newblock In \emph{Proceedings of the 2018 Conference on Empirical Methods in
  Natural Language Processing}, pages 875--880, Brussels, Belgium. Association
  for Computational Linguistics.

\bibitem[{Orife et~al.(2020)Orife, Kreutzer, Sibanda, Whitenack, Siminyu,
  Martinus, Ali, Abbott, Marivate, Kabongo, Meressa, Murhabazi, Ahia, van
  Biljon, Ramkilowan, Akinfaderin, Öktem, Akin, Kioko, Degila, Kamper, Dossou,
  Emezue, Ogueji, and Bashir}]{orife2020masakhane}
Iroro Orife, Julia Kreutzer, Blessing Sibanda, Daniel Whitenack, Kathleen
  Siminyu, Laura Martinus, Jamiil~Toure Ali, Jade Abbott, Vukosi Marivate,
  Salomon Kabongo, Musie Meressa, Espoir Murhabazi, Orevaoghene Ahia, Elan van
  Biljon, Arshath Ramkilowan, Adewale Akinfaderin, Alp Öktem, Wole Akin,
  Ghollah Kioko, Kevin Degila, Herman Kamper, Bonaventure Dossou, Chris Emezue,
  Kelechi Ogueji, and Abdallah Bashir. 2020.
\newblock Masakhane -- machine translation for africa.
\newblock \emph{arXiv preprint arXiv:2003.11529}.

\bibitem[{{\"O}stling and Tiedemann(2016)}]{Ostling2016efmaral}
Robert {\"O}stling and J{\"o}rg Tiedemann. 2016.
\newblock \href {http://ufal.mff.cuni.cz/pbml/106/art-ostling-tiedemann.pdf}
  {Efficient word alignment with {M}arkov {C}hain {M}onte {C}arlo}.
\newblock \emph{Prague Bulletin of Mathematical Linguistics}, 106:125--146.

\bibitem[{Ott et~al.(2019)Ott, Edunov, Baevski, Fan, Gross, Ng, Grangier, and
  Auli}]{ott-etal-2019-fairseq}
Myle Ott, Sergey Edunov, Alexei Baevski, Angela Fan, Sam Gross, Nathan Ng,
  David Grangier, and Michael Auli. 2019.
\newblock \href {https://doi.org/10.18653/v1/N19-4009} {fairseq: A fast,
  extensible toolkit for sequence modeling}.
\newblock In \emph{Proceedings of the 2019 Conference of the North {A}merican
  Chapter of the Association for Computational Linguistics (Demonstrations)},
  pages 48--53, Minneapolis, Minnesota. Association for Computational
  Linguistics.

\bibitem[{Ott et~al.(2018)Ott, Edunov, Grangier, and
  Auli}]{ott-etal-2018-scaling}
Myle Ott, Sergey Edunov, David Grangier, and Michael Auli. 2018.
\newblock \href {https://doi.org/10.18653/v1/W18-6301} {Scaling neural machine
  translation}.
\newblock In \emph{Proceedings of the Third Conference on Machine Translation:
  Research Papers}, pages 1--9, Brussels, Belgium. Association for
  Computational Linguistics.

\bibitem[{Papineni et~al.(2002)Papineni, Roukos, Ward, and
  Zhu}]{papineni-etal-2002-bleu}
Kishore Papineni, Salim Roukos, Todd Ward, and Wei-Jing Zhu. 2002.
\newblock \href {https://doi.org/10.3115/1073083.1073135} {{B}leu: a method for
  automatic evaluation of machine translation}.
\newblock In \emph{Proceedings of the 40th Annual Meeting of the Association
  for Computational Linguistics}, pages 311--318, Philadelphia, Pennsylvania,
  USA. Association for Computational Linguistics.

\bibitem[{Popovi{\'c}(2015)}]{popovic-2015-chrf}
Maja Popovi{\'c}. 2015.
\newblock \href {https://doi.org/10.18653/v1/W15-3049} {chr{F}: character
  n-gram {F}-score for automatic {MT} evaluation}.
\newblock In \emph{Proceedings of the Tenth Workshop on Statistical Machine
  Translation}, pages 392--395, Lisbon, Portugal. Association for Computational
  Linguistics.

\bibitem[{Post(2018)}]{post-2018-call}
Matt Post. 2018.
\newblock \href {https://doi.org/10.18653/v1/W18-6319} {A call for clarity in
  reporting {BLEU} scores}.
\newblock In \emph{Proceedings of the Third Conference on Machine Translation:
  Research Papers}, pages 186--191, Brussels, Belgium. Association for
  Computational Linguistics.

\bibitem[{Reid and Artetxe(2021)}]{reid2021paradise}
Machel Reid and Mikel Artetxe. 2021.
\newblock {PARADISE}: Exploiting parallel data for multilingual
  sequence-to-sequence pretraining.
\newblock \emph{ArXiv}, abs/2108.01887.

\bibitem[{Song et~al.(2019)Song, Tan, Qin, Lu, and Liu}]{song2019mass}
Kaitao Song, Xu~Tan, Tao Qin, Jianfeng Lu, and Tie{-}Yan Liu. 2019.
\newblock \href {http://proceedings.mlr.press/v97/song19d.html} {{MASS:} masked
  sequence to sequence pre-training for language generation}.
\newblock In \emph{Proceedings of the 36th International Conference on Machine
  Learning, {ICML} 2019, 9-15 June 2019, Long Beach, California, {USA}},
  volume~97 of \emph{Proceedings of Machine Learning Research}, pages
  5926--5936. {PMLR}.

\bibitem[{Sutskever et~al.(2014)Sutskever, Vinyals, and
  Le}]{sutskever2014equence}
Ilya Sutskever, Oriol Vinyals, and Quoc~V. Le. 2014.
\newblock \href
  {https://proceedings.neurips.cc/paper/2014/hash/a14ac55a4f27472c5d894ec1c3c743d2-Abstract.html}
  {Sequence to sequence learning with neural networks}.
\newblock In \emph{Advances in Neural Information Processing Systems 27: Annual
  Conference on Neural Information Processing Systems 2014, December 8-13 2014,
  Montreal, Quebec, Canada}, pages 3104--3112.

\bibitem[{Tiedemann(2012)}]{tiedemann-2012-parallel}
J{\"o}rg Tiedemann. 2012.
\newblock \href
  {http://www.lrec-conf.org/proceedings/lrec2012/pdf/463_Paper.pdf} {Parallel
  data, tools and interfaces in {OPUS}}.
\newblock In \emph{Proceedings of the Eighth International Conference on
  Language Resources and Evaluation ({LREC}'12)}, pages 2214--2218, Istanbul,
  Turkey. European Language Resources Association (ELRA).

\bibitem[{Tran et~al.(2020)Tran, Tang, Li, and Gu}]{tran2020crosslingual}
Chau Tran, Yuqing Tang, Xian Li, and Jiatao Gu. 2020.
\newblock Cross-lingual retrieval for iterative self-supervised training.
\newblock In \emph{Advances in Neural Information Processing Systems}.

\bibitem[{Vaswani et~al.(2017)Vaswani, Shazeer, Parmar, Uszkoreit, Jones,
  Gomez, Kaiser, and Polosukhin}]{vaswani2017attention}
Ashish Vaswani, Noam Shazeer, Niki Parmar, Jakob Uszkoreit, Llion Jones,
  Aidan~N. Gomez, Lukasz Kaiser, and Illia Polosukhin. 2017.
\newblock \href
  {https://proceedings.neurips.cc/paper/2017/hash/3f5ee243547dee91fbd053c1c4a845aa-Abstract.html}
  {Attention is all you need}.
\newblock In \emph{Advances in Neural Information Processing Systems 30: Annual
  Conference on Neural Information Processing Systems 2017, December 4-9, 2017,
  Long Beach, CA, {USA}}, pages 5998--6008.

\bibitem[{Wenzek et~al.(2020)Wenzek, Lachaux, Conneau, Chaudhary, Guzm{\'a}n,
  Joulin, and Grave}]{wenzek-etal-2020-ccnet}
Guillaume Wenzek, Marie-Anne Lachaux, Alexis Conneau, Vishrav Chaudhary,
  Francisco Guzm{\'a}n, Armand Joulin, and Edouard Grave. 2020.
\newblock \href {https://aclanthology.org/2020.lrec-1.494} {{CCN}et: Extracting
  high quality monolingual datasets from web crawl data}.
\newblock In \emph{Proceedings of the 12th Language Resources and Evaluation
  Conference}, pages 4003--4012, Marseille, France. European Language Resources
  Association.

\bibitem[{Xue et~al.(2020)Xue, Constant, Roberts, Kale, Al-Rfou, Siddhant,
  Barua, and Raffel}]{xue2020t5}
Linting Xue, Noah Constant, Adam Roberts, Mihir Kale, Rami Al-Rfou, Aditya
  Siddhant, Aditya Barua, and Colin Raffel. 2020.
\newblock mt5: A massively multilingual pre-trained text-to-text transformer.
\newblock \emph{arXiv preprint arXiv:2010.11934}.

\bibitem[{Zoph et~al.(2016)Zoph, Yuret, May, and
  Knight}]{zoph-etal-2016-transfer}
Barret Zoph, Deniz Yuret, Jonathan May, and Kevin Knight. 2016.
\newblock \href {https://doi.org/10.18653/v1/D16-1163} {Transfer learning for
  low-resource neural machine translation}.
\newblock In \emph{Proceedings of the 2016 Conference on Empirical Methods in
  Natural Language Processing}, pages 1568--1575, Austin, Texas. Association
  for Computational Linguistics.

\end{thebibliography}
\insertappendix
\end{document}